%% file: paper.tex
\documentclass{article}

\usepackage{microtype}
\usepackage{graphicx}
\usepackage{hyperref}
\usepackage{amsmath}
\usepackage{amssymb}
\usepackage{mathtools}
\usepackage{amsthm}
\usepackage{multirow}
\usepackage{pgfplots}

\usepackage[preprint]{neurips_2026}
\usepackage{common}

\title{Parallel Recursive LSTM}

\author{%
  Tristan Gaudreault \\
  School of Electrical Engineering and Computer Science\\
  University of Ottawa\\
  Ottawa, ON \\
  \texttt{tgaud061@uottawa.ca} \\
  \And
  Yongyi Mao \\
  School of Electrical Engineering and Computer Science \\
  University of Ottawa\\
  Ottawa, ON \\
  \texttt{ymao@uottawa.ca} \\
}

\begin{document}
\maketitle

\begin{abstract}
Transformers have become the dominant architecture for sequence modeling by using self-attention to enable expressive and highly parallel processing. However, the resulting quadratic time and memory costs limit efficiency in long-context settings. Recurrent models such as LSTMs provide explicit nonlinear state updates and strong state-tracking capabilities, yet their strictly sequential computation limits parallelism. We introduce the Parallel Recursive LSTM (PR-LSTM), a hierarchical recurrent architecture that replaces left-to-right recurrence with recursive nonlinear state composition over a balanced computation tree. Tokens are first mapped independently to latent states, which are then recursively merged by a learned gated composition block. This structure uses the reduction pattern underlying parallel scans as a fixed execution schedule, rather than assuming an associative recurrence. As a result, PR-LSTM retains nonlinear gated state representations while reducing recurrent parallel depth from linear to logarithmic. Empirically, PR-LSTM achieves strong sequence-length generalization on formal-language benchmarks, solving more tasks than standard RNN, LSTM, and Transformer baselines, while avoiding the quadratic scaling of attention. These results suggest that recurrent computation can be reorganized hierarchically to expose parallelism without restricting the transition dynamics to linear or associative forms. Code is available at \url{https://github.com/tristangaudreault/pr-lstm}.
\end{abstract}

\section{Introduction}
Modern deep learning systems have achieved unprecedented scale and capability, most prominently in the form of Transformer-based large language models (LLMs) \citep{NIPS2017_3f5ee243}. These architectures use self-attention to model pairwise interactions between tokens across the full context while enabling highly parallel computation. By stacking multiple layers, they construct rich hierarchical representations and capture long-range dependencies. This combination of expressive power and hardware efficiency has established Transformers as the dominant paradigm for modern sequence modeling.

Despite their empirical success, Transformers depart fundamentally from the stateful dynamics that characterized earlier recurrent approaches. Rather than maintaining a persistent hidden state that evolves over time, they repeatedly recompute representations at every layer. This design incurs quadratic computational and memory complexity in sequence length, limiting efficiency in long-context settings. More conceptually, attending to all inputs simultaneously contrasts with the sequential nature of human language processing and can be limiting for tasks that require stepwise reasoning or incremental information accumulation \citep{10.5555/3524938.3525416}. In addition, models with persistent hidden states provide an explicit object for analyzing how information evolves over time, a perspective used in prior work on visualizing and interpreting recurrent hidden state dynamics \citep{strobelt2017lstmvistoolvisualanalysis,ming2017understandinghiddenmemoriesrecurrent}.

Classical recurrent neural networks (RNNs) such as Long Short-Term Memory (LSTM) \citep{HochreiterSepp1997LSM} and Gated Recurrent Unit (GRU) \citep{cho2014learningphraserepresentationsusing} networks process input tokens sequentially, incrementally updating a fixed-size hidden state via nonlinear gating mechanisms. The sequential application incurs only linear computational complexity in sequence length, but inherently limits parallelization. Nonetheless, the sequential application remains beneficial for tasks that require state tracking \citep{deletang2023neural}.

Recent work has sought to reconcile the computational efficiency of state-space models with the parallelism of Transformers. A central mechanism underlying many of these approaches is the associative scan \citep{BlellochTR90}, which enables sequence processing in logarithmic depth. However, the algorithm requires an associative transition operator, making it incompatible with the nonlinear state dependencies of RNNs. One prominent response to this limitation is the development of structured state space models such as S4 \citep{gu2022efficiently}, which formulate sequence modeling through discretized linear dynamical systems with convolutional representations and structured parameterizations for efficient long-context processing. Subsequent variants, including S5 \citep{smith2023simplified}, DSS \citep{10.5555/3600270.3601940}, and Mamba \citep{gu2024mambalineartimesequencemodeling,10.5555/3692070.3692469}, further improve scalability through simplified parameterizations, input-dependent state transitions, and hardware-aware implementations that retain linear-time scaling while improving throughput and memory efficiency. In parallel, hybrid recurrent architectures such as RWKV \citep{peng2023rwkvreinventingrnnstransformer} and RetNet \citep{sun2023retentivenetworksuccessortransformer} reinterpret attention-like computation through recurrent or retention-based updates, seeking to approximate the contextual flexibility of attention while preserving recurrent-style inference efficiency. Other efficient sequence modeling approaches include linear attention mechanisms \citep{10.5555/3524938.3525416,choromanski2020rethinking}, which approximate softmax attention through kernel feature maps, reducing the quadratic complexity. Convolutional architectures such as Hyena \citep{poli2023hyenahierarchylargerconvolutional} instead employ implicit long convolutions and data-controlled gating to efficiently model long-range interactions. More recently, architectures such as xLSTM \citep{beck2024xlstmextendedlongshortterm} revisit classical gated recurrence at scale, demonstrating that carefully structured recurrent updates can remain competitive with modern sequence models. Despite their differing formulations, these methods generally obtain efficiency by imposing linear, simplified, or otherwise constrained transition dynamics, potentially limiting their ability to represent the rich nonlinear and context-dependent computations available to fully gated recurrent networks or unconstrained attention mechanisms \citep{merrill-sabharwal-2023-parallelism,10.5555/3692070.3693514}.

We introduce the Parallel Recursive LSTM (PR-LSTM), a hierarchical recurrent architecture that achieves logarithmic parallel depth through a work-efficient balanced reduction structure. Rather than processing information sequentially, PR-LSTM recursively composes neighboring state pairs using a learned LSTM encoder. Lower levels of the hierarchy capture local interactions, while higher levels progressively aggregate broader contextual information. PR-LSTM adopts the reduction structure of associative scans purely as a parallel execution strategy for a learned non-associative composition operator. In practice, this pattern can be implemented with scan primitives that accept user-defined binary operators, such as \texttt{jax.lax.associative\_scan} in JAX, provided that the result is interpreted as a fixed balanced-tree computation rather than as an associative prefix computation. Empirically, PR-LSTM shows strong formal-language generalization, succeeding on more tasks than both Transformer and sequential LSTM baselines under the protocol of \citet{deletang2023neural}. These results suggest that PR-LSTM retains key recurrent capabilities, including counting and long-term state tracking. It also exposes substantially more parallelism than standard recurrent models, while preserving linear total work in sequence length and avoiding the quadratic activation growth of attention. Together, these findings position PR-LSTM as a middle ground between strictly sequential recurrence and fully attention-based sequence modeling. While our experiments focus on formal-language generalization, they provide evidence that nonlinear recurrent computation can be reorganized to better exploit parallel hardware. 

Our contributions are:
\begin{itemize}
    \item We introduce PR-LSTM, a recurrent architecture that restructures LSTM-style computation into a balanced hierarchy of state updates, rather than a strictly sequential chain.
    \item We propose a multi-stage LSTM-based encoder that merges neighboring latent states through nonlinear gated updates, enabling recurrent computation over a balanced hierarchy.
    \item We use the parallel scan reduction pattern as an execution schedule, exposing parallelism without assuming an associative recurrence.
    \item We show that PR-LSTM achieves strong formal-language length generalization while reducing recurrent parallel depth and avoiding the quadratic scaling of attention.
\end{itemize}

\section{Related work}

\subsection{Parallel sequence models}

Many efficient sequence models achieve parallelism by imposing structure on the state transition. In structured state-space models and related recurrent architectures, linear or associative update rules make parallel scan algorithms applicable, enabling logarithmic-depth sequence processing. However, these constraints can limit the range of sequence-processing tasks that such models can reliably solve \citep{merrill-sabharwal-2023-parallelism,10.5555/3692070.3693514}. Rather than requiring the transition rule itself to be associative, PR-LSTM moves the source of parallelism from the transition rule to the computation graph.

A separate line of work parallelizes nonlinear recurrence through iterative solvers. Recent approaches \citep{danieli2025pararnnunlockingparalleltraining,lim2024parallelizing,gonzalez2024towards} cast RNN evaluation as a global nonlinear system and solve it in parallel using Newton iterations. These methods enable parallel sequence evaluation, but typically depend on convergence assumptions and structured approximations that avoid forming full Jacobians. In contrast, PR-LSTM defines parallel nonlinear recurrence directly through hierarchical state composition rather than iterative global optimization. Parallelism is therefore built into the architecture while retaining nonlinear gated state transitions.

\subsection{Hierarchical recurrence}

Recursive Neural Networks \citep{10.5555/3104482.3104499} process hierarchical structures such as syntactic parse trees. Tree-LSTM architectures \citep{tai2015improvedsemanticrepresentationstreestructured} extend this idea with gated memory updates. In these models, the recursive computation graph is typically provided by an external linguistic or semantic hierarchy. PR-LSTM instead constructs its hierarchy for parallel execution, using the work-efficient balanced reduction pattern of associative scans as its computation graph.

Our recursive composition block is inspired by the binary Tree-LSTM cell, but adapted for parallel sequence composition. Input-dependent dynamics are delegated to a separate state embedding module, allowing the recursive block to focus on latent state merging and transformation. After each binary composition, PR-LSTM applies a refinement stage, yielding a structure that echoes the multi-stage organization of Transformer encoder blocks.

\section{Approach}\label{sec:approach}

In this work, the dimensionality of a vector $\bv v$ is denoted by $d_v$. Let an input sequence be written as $[\bv x_1, \bv x_2, \dots, \bv x_T],\,\bv x_t \in \mathbb{R}^{d_x}$ where $T \in \mathbb{N}^+$ denotes the sequence length. 

\subsection{Parallel Recursive LSTM}

\begin{figure}[t]
    \centering
    \begin{subfigure}[t]{0.46\columnwidth}
    \centering
    \includegraphics[height=6.5cm]{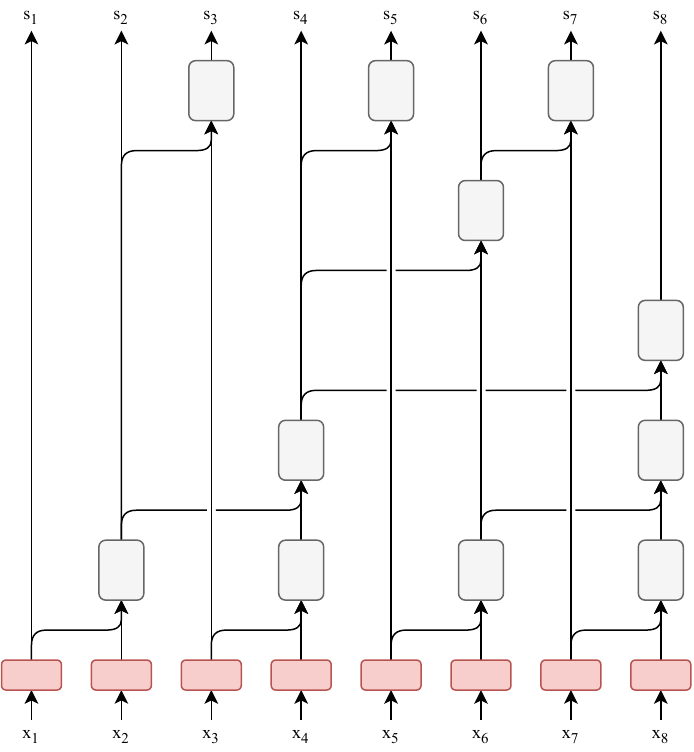}
    \caption{Embedded token states are recursively composed over a balanced scan-style graph, yielding logarithmic parallel depth. Red cells denote the state embedding module, while grey cells denote the nonlinear LSTM encoder.}
    \label{fig:tree}
    \end{subfigure}
    \hfill
    \begin{subfigure}[t]{0.46\columnwidth}
    \centering
    \includegraphics[height=6.5cm]{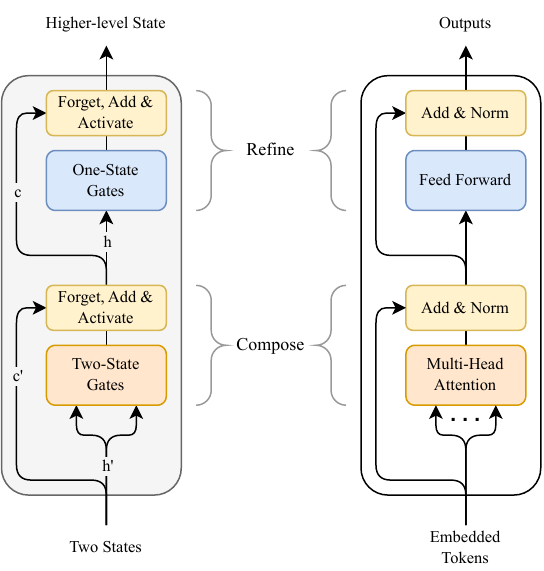}
    \caption{The LSTM encoder (left) performs binary state composition followed by unary refinement, loosely paralleling the two-stage organization of a Transformer encoder (right).}
    \label{fig:encoders}
    \end{subfigure}
    \caption{\textbf{Overview of PR-LSTM.}}
    \label{fig:overview}
\end{figure}

The Parallel Recursive LSTM (PR-LSTM) is a hierarchical recurrent architecture that replaces strictly sequential recurrence with recursive state composition over a work-efficient balanced computation graph. Input tokens are first mapped independently to latent states by a state embedding module, allowing this stage to be computed fully in parallel across the sequence. The proposed LSTM encoder is then applied recursively to neighboring state pairs according to the balanced reduction schedule illustrated in Figure~\ref{fig:tree}. Although the LSTM encoder is not assumed to be associative, we use the scan reduction pattern only as a fixed execution schedule, interpreting the result as a particular balanced-tree computation rather than as an associative prefix computation. In practice, this pattern can be implemented with scan-style primitives that accept user-defined binary operators, such as \texttt{jax.lax.associative\_scan} in JAX, or analogous framework-specific implementations.

Rather than reproducing the exact dynamics of a sequential RNN, PR-LSTM learns hierarchical state compositions over the input sequence. Lower levels of the tree capture local interactions between neighboring representations, while higher levels progressively aggregate broader contextual information. The resulting representations are therefore shaped jointly by the recursive computation graph and the nonlinear gating dynamics of the encoder. In addition, the balanced tree structure shortens the effective communication path between distant tokens. In a sequential RNN, information from two positions separated by distance $t$ must propagate through $O(t)$ recurrent transitions. In PR-LSTM, the recursive computation reduces this path length to $O(\log t)$, potentially improving long-range information flow.

\subsection{LSTM encoder}

The LSTM encoder is the nonlinear composition operator applied at each internal node of the recursive computation structure. Each state is a pair $(\bv h, \bv c)$, where $\bv h \in \mathbb{R}^{d_h}$ denotes the hidden state and $\bv c \in \mathbb{R}^{d_h}$ denotes the cell state. As shown in Figure~\ref{fig:encoders}, the encoder combines two incoming states into a higher-level state through binary composition followed by unary refinement. The corresponding update equations are given in Figure~\ref{fig:definitions}.

The composition stage computes gates from the two incoming hidden states and applies a forget/add/activate update. Forget gates modulate the two cell states, while input and update terms introduce new content derived from the interaction between hidden states.

The refinement stage further transforms the composed state using single-state gating and another forget/add/activate update. This stage can be stacked to increase nonlinear depth, or omitted for a leaner architecture.

The design is loosely inspired by Transformer encoder blocks, as reflected by the color correspondence in Figure~\ref{fig:encoders}. However, the two architectures aggregate information differently. Self-attention mixes information over a variable-size context in a single layer, whereas the LSTM encoder combines only a fixed number of states at a time and therefore aggregates sequence-level information recursively through the computation tree.\footnote{This work focuses on the two-input configuration. More generally, the encoder can be defined for larger fixed arities, yielding recursive computation trees with higher branching factors.}

\begin{figure}[h]
\input{figures/input/encoder_equations}
\caption{\textbf{Definitions of the color-coded LSTM encoder modules used in Figure~\ref{fig:overview}.} $\Linear_d(\cdot)$ denotes a learned affine map with output dimension $d$, different occurrences do not share parameters.}
\label{fig:definitions}
\end{figure}

\subsection{Computational properties}\label{sec:parallel}

We analyze parallel runtime using Brent's theorem \citep{Gustafson2011}. Given an algorithm's total work $\tau_1$ and parallel depth $\tau_N$, its runtime on $p$ processors satisfies
\begin{equation}
    \tau_p = O\left(\tau_N + \frac{\tau_1}{p}\right).
\end{equation}

\begin{table}[t]
    \centering
    \caption{\textbf{Growth rates of parallel computation time.} We consider input length 
    $T \in \mathbb{N}^+$ and number of available processors $p \in \mathbb{N}^+$.}
    \subimport{figures/tabular}{parallel}
    \label{tab:parallel}
\end{table}

Intuitively, runtime is depth-limited before processor resources are saturated, and work-limited thereafter. For long sequences under a limited processor budget, this bound favors PR-LSTM, as it preserves the linear work of an RNN while reducing recurrent depth from $O(T)$ to $O(\log T)$. The same structural distinction is relevant for memory. Because PR-LSTM composes states over a balanced reduction graph rather than through all-pairs token interactions, it avoids the quadratic activation growth associated with attention-based sequence mixing.

\subsection{Limitations}\label{sec:limitations}

PR-LSTM also introduces several limitations. First, the balanced recursive structure imposes a fixed hierarchical inductive bias that may not align with the dependency structure of every sequence modeling task. Unlike attention mechanisms, which dynamically route information between arbitrary tokens, PR-LSTM aggregates information through a predetermined reduction topology. As a result, tasks with strongly sequential or order-sensitive dynamics may require the model to preserve and propagate an increasingly large set of latent possibilities throughout the recursive aggregation process.

Second, although the architecture exposes substantial parallelism, it also introduces overhead relative to strictly sequential recurrence. The recursive reduction adds internal composition operations and implementation complexity, even while reducing the parallel depth to logarithmic. Moreover, the proposed LSTM encoder uses $14$ weight matrices, so its parameter count scales quadratically with the hidden dimension with a larger constant factor than the $4$ matrices used in a conventional LSTM cell. Reconstructing token-level contextual states from the recursive hierarchy may also introduce additional implementation and memory-management complexity relative to standard recurrent architectures.

Finally, our empirical evaluation is limited to formal-language tasks. While these settings isolate key aspects of sequence processing and compositional reasoning, evaluating PR-LSTM on large-scale language modeling, multimodal learning, and long-context reasoning benchmarks remains future work. These results should therefore be interpreted as evidence about controlled length generalization and state-tracking behavior, rather than as direct evidence of competitiveness on open-ended language modeling.

\section{Experiments}\label{sec:experiments}

We evaluate PR-LSTM on the formal-language benchmark of \citet{deletang2023neural}, which compares sequence models across $15$ tasks: $4$ regular languages (R), $4$ deterministic context-free languages (DCF), and $7$ context-sensitive languages (CS). Unless otherwise stated, we follow the experimental protocol and hyperparameter settings of the original study. We state all deviations from this setup in the corresponding experimental descriptions.

\subsection{Main results}\label{sec:generalization}
\begin{table*}[]
    \centering
    \caption{\textbf{Main Results}. Score is accuracy averaged over test lengths $T\in\{41,42,\dots, 500\}$, maximized over 10 seeds. Generalization is \textbf{successful} if $\text{score} \ge 90\%$. Baseline results from \citep{deletang2023neural}, we report new results for PR-LSTM. Grey shading highlights our method. SRNN denotes Stack-RNN, TRNN denotes Tape-RNN, and Transformer refers to the encoder-only Transformer baseline.}
    \subimport{figures/tabular}{big_table}
    \label{tab:big-table}
\end{table*}

We first evaluate whether hierarchical nonlinear recurrence preserves the long-range state-tracking and compositional generalization capabilities typically associated with recurrent architectures. Models are trained on sequence lengths $T \sim \operatorname{Uniform}(\{1,2,\dots,40\})$ and evaluated on longer sequences with $T \in \{41,42,\dots,500\}$. Following \citet{deletang2023neural}, performance is measured by averaging accuracy across all test lengths, and a model is considered to successfully generalize if $\text{score} \ge 90\%$.

As shown in Table~\ref{tab:big-table}, PR-LSTM successfully generalizes on more tasks ($6/15$) than any non-memory-augmented baseline, including LSTM ($5/15$), RNN ($4/15$), and Transformer ($2/15$). These results suggest that balanced hierarchical recurrence can expose substantially more parallel execution than strictly sequential recurrence while retaining state-based mechanisms that support length generalization in the evaluated formal-language setting.

At the task level, PR-LSTM solves Missing Duplicate, a task otherwise only solved by the memory-augmented Tape RNN. This suggests that the recursive hierarchy can support at least some forms of long-range information retrieval. PR-LSTM also preserves the counting behavior required for Bucket Sort and the finite-state tracking needed for regular grammar tasks. Its main weaknesses appear on Modular Arithmetic and Solve Equation, where it generalizes substantially worse than the LSTM and RNN baselines, although it still outperforms the Transformer encoder.

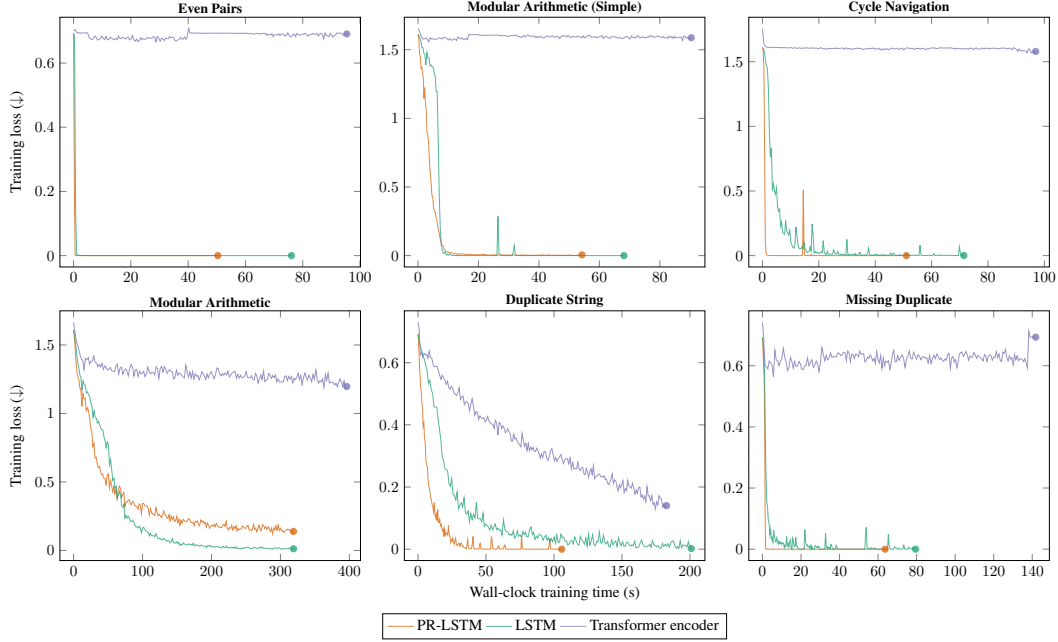
\begin{figure}
    \centering
    \resizebox{\linewidth}{!}{\subimport{figures/tikz}{time_efficiency}}
    \caption{\textbf{Training dynamics across six tasks}. Training loss (y-axis) as a function of wall-clock training time (x-axis), with all runs trained for $40{,}000$ iterations on a cluster job allocation with an NVIDIA H100, 1 CPU core, and 35 GB of system memory. Endpoint markers show completion times for this fixed training budget, highlighting relative training speed.}
    \label{fig:time-efficiency}
\end{figure}

Beyond generalization performance, we evaluate whether the hierarchical computation structure yields practical efficiency gains during training. Figure~\ref{fig:time-efficiency} compares training loss as a function of wall-clock time across six tasks, with full results provided in Appendix~\ref{app:full_results}. The endpoint of each curve, marked with a dot, indicates the total time required to train for $40{,}000$ iterations.

Across most tasks, PR-LSTM reaches low training loss faster than the LSTM and Transformer encoder baselines, while also taking less total training time. The Transformer encoder follows the original experimental setup, with $5$ layers, each containing one attention layer and two feed-forward layers. As a result, it is a substantially heavier baseline in this setting, which is reflected in its slower training. These results indicate that the parallelism exposed by the recursive computation structure translates into practical wall-clock gains in this benchmark setting.

\begin{figure}
    \centering
    \begin{subfigure}{0.49\textwidth}
        \centering
        \resizebox{!}{7.75cm}{\subimport{figures/tikz}{profiling_h100}}
        \caption{Cluster allocation with NVIDIA H100.}
    \end{subfigure}
    \hfill
    \begin{subfigure}{0.49\textwidth}
        \centering
        \resizebox{!}{7.75cm}{\subimport{figures/tikz}{profiling_rtx4070}}
        \caption{Local workstation with NVIDIA RTX 4070.}
    \end{subfigure}
    \caption{\textbf{Scaling behavior of inference time and memory consumption with increasing sequence length.} Colors follow Figure~\ref{fig:time-efficiency}. Results are averaged over $100$ batches, each containing $1024$ samples. A $\triangle$ marks termination due to exceeding the inference-time threshold $\tau_p \ge 500\text{ms}$, while a $\square$ marks termination due to GPU memory exhaustion. Measurements were collected on two hardware environments: a cluster job allocation with an NVIDIA H100, 1 CPU core, and 35 GB of system memory, and a local workstation with an NVIDIA RTX 4070, 20 CPU cores, and 32 GB of system memory.}
    \label{fig:time-growth}
\end{figure}
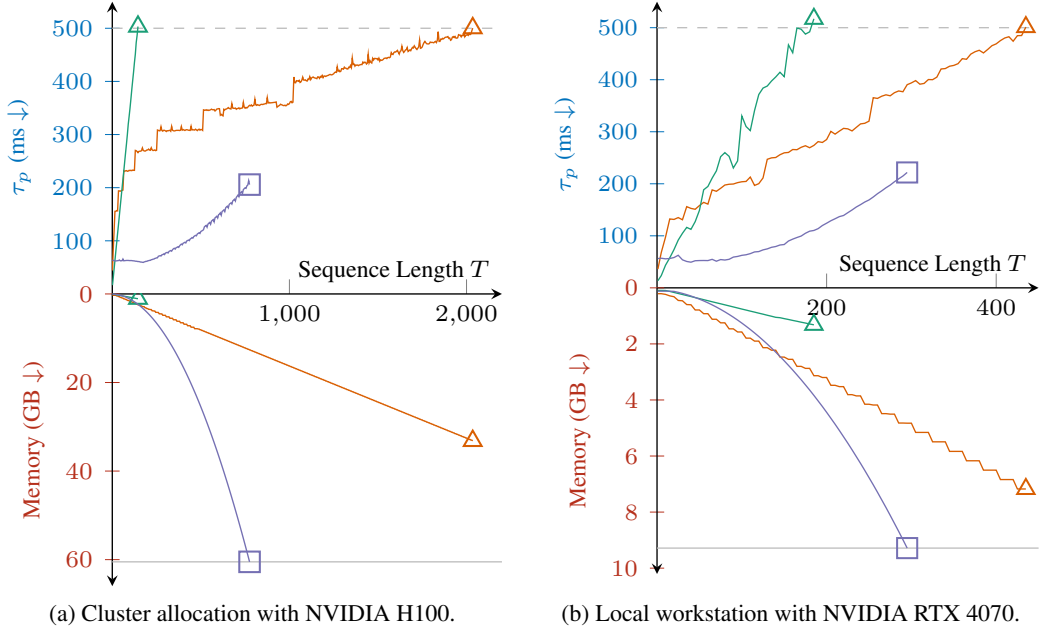

Finally, Figure~\ref{fig:time-growth} evaluates the scaling behavior described in Section~\ref{sec:parallel}. We measure average inference time and memory consumption while increasing sequence length. Results are averaged over 100 batches, each containing 1024 samples. Each curve terminates when the corresponding model either exceeds the inference-time threshold $\tau_p \ge 500\text{ms}$, marked with a triangle, or exhausts available memory, marked with a square.

Across both the NVIDIA H100 and RTX 4070 environments, PR-LSTM exhibits substantially lower inference-time growth than the sequential LSTM baseline, consistent with its logarithmic recurrent depth. Its memory consumption grows approximately linearly with sequence length, as expected from its linear total work. In contrast, the Transformer encoder initially benefits from parallel execution, but its all-pairs attention mechanism leads to faster growth in both runtime and memory consumption. Consequently, the Transformer reaches the memory limit at shorter sequence lengths, whereas the LSTM and PR-LSTM runs terminate by exceeding the inference-time threshold.

Taken together, these results suggest that hierarchical nonlinear recurrence can preserve important recurrent computational behaviors while exposing substantially greater parallelism than strictly sequential recurrent architectures.

\subsection{Ablations}\label{sec:ablations}

\begin{table*}[]
    \centering
    \caption{\textbf{Ablation Results}. Score denotes the accuracy averaged over test lengths $T\in\{41,42,\dots, 500\}$. Generalization is considered \textbf{successful} if $\text{score} \ge 90\%$. Our main architecture is shaded. The number of refining stages is denoted as $R$.}
    \subimport{figures/tabular}{big_table_ablation}
    \label{tab:ablation}
\end{table*}
We perform ablations to isolate which components of PR-LSTM contribute to its generalization behavior. In particular, we examine whether the gains are explained by increased parameter count, whether additional refinement depth improves recursive composition, and whether gated state updates are necessary for stable hierarchical recurrence. Across all variants, the balanced recursive computation graph is kept fixed, allowing us to separate the effect of the hierarchy itself from the state-processing mechanism used at each node.

\paragraph{Parameter count.}
The proposed PR-LSTM encoder has more parameters than a conventional LSTM cell, since it uses $14$ hidden-dimension-scaling weight matrices instead of $4$. To examine whether the observed gains are solely due to this increased parameter budget, we construct a parameter-matched variant by reducing the hidden dimension by a factor of $\sqrt{14/4}$. In our experimental setup, this changes the hidden size from $256$ to $137$, yielding an approximate match to the parameter count of the LSTM baseline. As shown in Table~\ref{tab:ablation}, the parameter-matched PR-LSTM preserves similar generalization behavior to the full model, solving the same number of tasks and retaining the distinctive success on Missing Duplicate. This provides evidence that the recursive hierarchy contributes to these behaviors beyond the effect of increased parameter count.

\paragraph{Encoder depth.}
We next evaluate the contribution of refinement depth by comparing the baseline model against variants with $0$ and $2$ refining stages. As shown in Table~\ref{tab:ablation}, the $0$ stage variant fails to generalize on several tasks where the deeper variants succeed, including Modular Arithmetic (Simple), Cycle Navigation, and Missing Duplicate. This suggests that additional nonlinear processing after each binary composition helps the model form useful higher-level states, rather than merely passing forward a shallow pairwise merge. However, increasing the depth beyond the baseline does not yield consistent improvements across tasks. We therefore use the baseline depth as a tradeoff between expressivity and parameter efficiency.

\paragraph{State-processing architecture.}
Finally, we test whether the LSTM-style gating mechanism is essential, or whether the recursive hierarchy alone is sufficient. We replace the LSTM-based encoder with stacked $\relu$ transformations, defining both composition and refinement stages as
\begin{equation}
    \operatorname{FC}(\bv h) \coloneq \relu(\bv W \bv h + \bv b).
\end{equation}
In this PR-RNN variant, the cell state is removed and each node carries only a hidden state vector. Binary composition concatenates the two incoming hidden states before applying the ReLU transformation, while refinement applies the same form to a single hidden state. This keeps the recursive computation graph fixed while isolating the effect of removing gated memory dynamics. As shown in Table~\ref{tab:ablation}, these variants generalize on substantially fewer tasks than their depth-matched PR-LSTM counterparts. The degradation is most pronounced on tasks requiring long-range state tracking and compositional reasoning. For example, PR-RNN variants fail to match PR-LSTM on Bucket Sort. The addition of refining stages improves performance on Missing Duplicate, but collapses performance on Modular Arithmetic (Simple). These results suggest that the recursive hierarchy benefits from an explicit gated memory state rather than hidden state transformations alone.

\section{Conclusion}
We introduced PR-LSTM, a hierarchical recurrent architecture that replaces the strictly sequential chain of LSTM updates with recursive composition over a balanced computation tree. This design exposes logarithmic parallel depth for nonlinear gated transitions. On formal-language tasks, PR-LSTM retains several important recurrent capabilities, including finite-state tracking, counting, and long-range retrieval, while improving parallel depth relative to standard sequential recurrence. These results suggest that nonlinear recurrence can remain a promising foundation for scalable sequence modeling when its computation is reorganized hierarchically.

\raggedbottom
\bibliographystyle{dinat}
\bibliography{refs}
\appendix

\section{Full results}\label{app:full_results}
\begin{figure}[H]
    \centering
    \resizebox{\linewidth}{!}{\subimport{figures/tikz}{time_efficiency_full}}
    \caption{\textbf{Training dynamics across all fifteen tasks}. Training loss (y-axis) as a function of wall-clock training time (x-axis), with all runs trained for $40{,}000$ iterations on a job allocation with an NVIDIA H100, 1 CPU core, and 35 GB of system memory.}
    \label{fig:time-efficiency_full}
\end{figure}
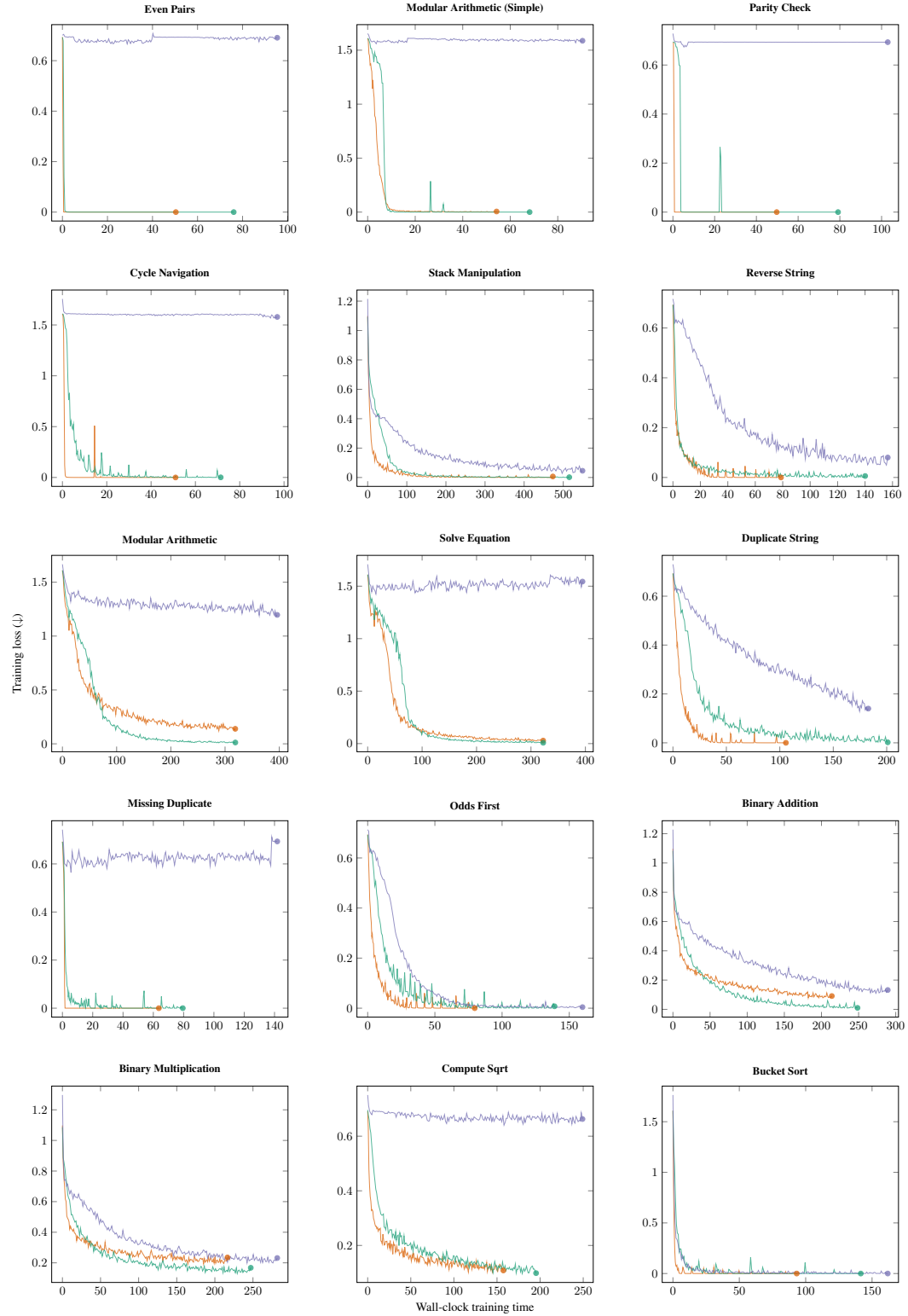
\end{document}

%% file: figures/input/encoder_equations.tex
\begin{minipage}[t]{0.49\textwidth}
\vspace{0pt}
\begin{tcolorbox}[
center title,
title={Two-State Gates},
colframe=AFrame,
colbacktitle=ABack,
colback=white,
coltitle=black,
fonttitle=\bfseries,
left=2mm,
right=2mm
]
Given $\bv h' \in \mathbb{R}^{2d_h}$, return
\begin{align}
    [\bv f_1, \bv f_2, \bv i, \bv o]
        &\coloneq \sigma \left(\Linear_{4d_h}(\bv h')\right), \\
    \bv u
        &\coloneq \tanh \left(\Linear_{d_h}(\bv h')\right).
\end{align}
\end{tcolorbox}
\vspace{-5pt}
\begin{tcolorbox}[
center title,
title={Forget, Add \& Activate},
colframe=BFrame,
colbacktitle=BBack,
colback=white,
coltitle=black,
fonttitle=\bfseries,
left=2mm,
right=2mm
]
Given $\{(\bv f_j, \bv c_j)\}_{j=1}^n$ with $\bv f_j, \bv c_j \in \mathbb{R}^{d_h}$ and $\bv i, \bv u, \bv o \in \mathbb{R}^{d_h}$, return
\begin{align}
    \bv c &\coloneq \bv i \odot \bv u + \sum_{j=1}^n \bv f_j \odot \bv c_j, \\
    \bv h &\coloneq \bv o \odot \tanh (\bv c).
\end{align}
\end{tcolorbox}
\end{minipage}
\hfill
\begin{minipage}[t]{0.49\textwidth}
\vspace{0pt}
\begin{tcolorbox}[
center title,
title={One-State Gates},
colframe=CFrame,
colbacktitle=CBack,
colback=white,
coltitle=black,
fonttitle=\bfseries,
left=2mm,
right=2mm
]
Given $\bv h \in \mathbb{R}^{d_h}$, return
\begin{align}
    [\bv f_1, \bv i, \bv o]
        &\coloneq \sigma \left(\Linear_{3d_h}(\bv h)\right), \\
    \bv u
        &\coloneq \tanh \left(\Linear_{d_h}(\bv h)\right).
\end{align}
\end{tcolorbox}
\vspace{-5pt}
\begin{tcolorbox}[
center title,
title={State Embedding},
colframe=DFrame,
colbacktitle=DBack,
colback=white,
coltitle=black,
fonttitle=\bfseries,
left=2mm,
right=2mm
]
Given $\bv x \in \mathbb{R}^{d_x}$, compute
\begin{align}
    [\bv i, \bv o] &\coloneq \sigma \left(\Linear_{2d_h}(\bv x)\right), \\
    \bv u &\coloneq \tanh \left(\Linear_{d_h}(\bv x)\right),
\end{align}
and return
\begin{align}
    \bv c &\coloneq \bv i \odot \bv u, \\
    \bv h &\coloneq \bv o \odot \tanh (\bv c) \label{eq:last}.
\end{align}
\end{tcolorbox}
\end{minipage}

%% file: figures/tabular/parallel.tex
\renewcommand{\arraystretch}{2.0}
\begin{tabular}{c|ccc}
    Model & RNN & PR-LSTM & Transformer \\
    \hline
    $O(\tau_p)$ & $T$ & $\log T + \frac{T}{p}$ & $1 + \frac{T^2}{p}$ \\
\end{tabular}

%% file: figures/tabular/big_table.tex
\renewcommand{\arraystretch}{1.2}
\begin{tabular}{|cc|ccccc|>{\columncolor{gray!10}}c|}
    \hline
    \multirow{2}{*}{Level} & \multirow{2}{*}{Task} & \multicolumn{6}{c|}{Score (\%)} \\
    \hhline{~~------}
    & & RNN & SRNN & TRNN & {\small Transformer} & LSTM & PR-LSTM \\
    \hline
    \multirow{4}{*}{R} & %
    Even Pairs & \textbf{100} & \textbf{100} & \textbf{100} & \textbf{96.4} & \textbf{100} & \textbf{100}  \\
    & \makecell{Modular Arithmetic\\(Simple)} & \textbf{100} & \textbf{100} & \textbf{100} & 24.2 & \textbf{100} & \textbf{99.7} \\
    & Parity Check & \textbf{100} & \textbf{100} & \textbf{100} & 52.0 & \textbf{100} & \textbf{100} \\
    & Cycle Navigation & \textbf{100} & \textbf{100} & \textbf{100} & 61.9 & \textbf{100} & \textbf{100} \\
    \hline
    \multirow{4}{*}{DCF} & %
    Stack Manipulation & 56.0 & \textbf{100} & \textbf{100} & 57.5 & 59.1 & 56.3 \\
    & Reverse String & 62.0 & \textbf{100} & \textbf{100} & 62.3 & 60.9 & 55.9 \\
    & Modular Arithmetic & 41.3 & \textbf{96.1} & \textbf{95.4} & 32.5 & 59.2 & 36.4 \\
    & Solve Equation & 51.0 & 56.2 & 64.4 & 25.7 & 67.8 & 33.7 \\
    \hline
    \multirow{4}{*}{CS} & %
    Duplicate String & 50.3 & 52.8 & \textbf{100} & 52.8 & 57.6 & 54.7 \\
    & Missing Duplicate & 52.3 & 55.2 & \textbf{100} & 56.4 & 54.3 & \textbf{100} \\
    & Odds First & 51.0 & 51.9 & \textbf{100} & 52.8 & 55.6 & 55.0 \\
    & Binary Addition & 50.3 & 52.7 & \textbf{100} & 54.3 & 55.5 & 51.8 \\
    & Binary Multiplication & 50.0 & 52.7 & 58.5 & 52.2 & 53.1 & 52.9 \\
    & Compute Sqrt & 54.3 & 56.5 & 57.8 & 52.4 & 57.5 & 56.8 \\
    & Bucket Sort & 27.9 & 78.1 & 70.7 & \textbf{91.9} & \textbf{99.3} & \textbf{99.4} \\
    \hline
    \multicolumn{2}{|c|}{Tasks Solved} & 4 & 7 & 11 & 2 & 5 & 6 \\
    \hline
\end{tabular}

%% file: figures/tikz/time_efficiency.tex
\begin{tikzpicture}
    \begin{groupplot}[
        group style={
            group size=3 by 2,
            horizontal sep=1cm,
            vertical sep=1cm,
        },
        title style={
            yshift=-0.25cm,
            font=\small\bf,
        },
        enlarge x limits=0.05,
        enlarge y limits=0.05,
        cycle list name=primary,
        every axis plot/.style={opacity=0.75}
    ]
    \nextgroupplot[title={Even Pairs}, legend to name=grouplegend, legend columns=-1, ylabel={Training loss ($\downarrow$)}]
    \addplot+[mark=*, mark indices={201}] table [x=time, y=loss] {figures/dat/even_pairs-lscm-training.dat}; 
    \addlegendentry{PR-LSTM}
    \addplot+[mark=*, mark indices={201}] table [x=time, y=loss] {figures/dat/even_pairs-lstm-training.dat}; 
    \addlegendentry{LSTM}
    \addplot+[mark=*, mark indices={201}] table [x=time, y=loss] {figures/dat/even_pairs-transformer_encoder-training.dat}; 
    \addlegendentry{Transformer encoder}
    \nextgroupplot[title={Modular Arithmetic (Simple)}]
    \addplot+[mark=*, mark indices={201}] table [x=time, y=loss] {figures/dat/modular_arithmetic-lscm-training.dat};
    \addplot+[mark=*, mark indices={201}] table [x=time, y=loss] {figures/dat/modular_arithmetic-lstm-training.dat};
    \addplot+[mark=*, mark indices={201}] table [x=time, y=loss] {figures/dat/modular_arithmetic-transformer_encoder-training.dat};
    \nextgroupplot[title={Cycle Navigation}]
    \addplot+[mark=*, mark indices={201}] table [x=time, y=loss] {figures/dat/cycle_navigation-lscm-training.dat};
    \addplot+[mark=*, mark indices={201}] table [x=time, y=loss] {figures/dat/cycle_navigation-lstm-training.dat};
    \addplot+[mark=*, mark indices={201}] table [x=time, y=loss] {figures/dat/cycle_navigation-transformer_encoder-training.dat};
    \nextgroupplot[title={Modular Arithmetic}, ylabel={Training loss ($\downarrow$)}]
    \addplot+[mark=*, mark indices={201}] table [x=time, y=loss] {figures/dat/modular_arithmetic_brackets-lscm-training.dat};
    \addplot+[mark=*, mark indices={201}] table [x=time, y=loss] {figures/dat/modular_arithmetic_brackets-lstm-training.dat};
    \addplot+[mark=*, mark indices={201}] table [x=time, y=loss] {figures/dat/modular_arithmetic_brackets-transformer_encoder-training.dat};
    \nextgroupplot[title={Duplicate String}, xlabel={Wall-clock training time (s)}]
    \addplot+[mark=*, mark indices={201}] table [x=time, y=loss] {figures/dat/duplicate_string-lscm-training.dat};
    \addplot+[mark=*, mark indices={201}] table [x=time, y=loss] {figures/dat/duplicate_string-lstm-training.dat};
    \addplot+[mark=*, mark indices={201}] table [x=time, y=loss] {figures/dat/duplicate_string-transformer_encoder-training.dat};
    \nextgroupplot[title={Missing Duplicate}]
    \addplot+[mark=*, mark indices={201}] table [x=time, y=loss] {figures/dat/missing_duplicate_string-lscm-training.dat};
    \addplot+[mark=*, mark indices={201}] table [x=time, y=loss] {figures/dat/missing_duplicate_string-lstm-training.dat};
    \addplot+[mark=*, mark indices={201}] table [x=time, y=loss] {figures/dat/missing_duplicate_string-transformer_encoder-training.dat};
    \end{groupplot}
    \node at (group c2r2.south) [yshift=-1.5cm] {\pgfplotslegendfromname{grouplegend}};
\end{tikzpicture}

%% file: figures/tikz/profiling_h100.tex
\begin{tikzpicture}
\begin{axis}[
    axis x line=middle,
    axis y line=left,
    xlabel={Sequence Length $T$},
    ylabel={$\tau_p$ (ms $\downarrow$)},
    xmin=0, xmax=2200,
    ymin=0, ymax=550,
    ylabel style={color=RoyalBlue},
    xtick align=inside,
    xtick={1000,2000},
    ytick={100, 200, 300, 400, 500},
    ytick style={color=RoyalBlue},
    ylabel style={yshift=-5pt},
    yticklabel style={color=RoyalBlue},
    cycle list name=primary,
    extra y ticks = {500},
    extra y tick style = {
        grid = major,
        major grid style = {dashed},
    tick style = {draw=none}
    },
    extra y tick label = \empty,
    width=4cm,
    height=3cm,
    scale only axis,
    font=\scriptsize
]
\addplot table [x=length, y=time] {figures/dat/parity_check-ntr-speed-1024-full.dat} node[pos=1, opacity=1.0, scale=1.5, solid] {\pgfuseplotmark{triangle}};
\addplot table [x=length, y=time] {figures/dat/parity_check-lstm-speed-1024-full.dat} node[pos=1, opacity=1.0, scale=1.5, solid] {\pgfuseplotmark{triangle}};
\addplot table [x=length, y=time] {figures/dat/parity_check-transformer_encoder-speed-1024-full.dat} node[pos=1, opacity=1.0, scale=1.5, solid] {\pgfuseplotmark{square}};
\end{axis}

\begin{axis}[
    at={(0,0)},
    anchor=origin,
    axis x line=none,
    axis y line=left,
    ylabel={Memory (GB $\downarrow$)},
    xmin=0, xmax=2200,
    ymin=0, ymax=66,
    ylabel style={color=BrickRed},
    ytick style={color=BrickRed},
    xtick={1000,2000},
    yticklabel style={color=BrickRed},
    ylabel style={yshift=-5pt},
    cycle list name=primary,
    y dir=reverse,
    extra y ticks = {60.5},
    extra y tick style = {
        grid = major,
    tick style = {draw=none}
    },
    extra y tick label = \empty,
    width=4cm,
    height=3cm,
    scale only axis,
    font=\scriptsize
]
\addplot table [x=length, y={peak_mem}] {figures/dat/parity_check-ntr-speed-1024-full.dat} node[pos=1, opacity=1.0, scale=1.5, solid] {\pgfuseplotmark{triangle}};
\addplot table [x=length, y={peak_mem}] {figures/dat/parity_check-lstm-speed-1024-full.dat} node[pos=1, opacity=1.0, scale=1.5, solid] {\pgfuseplotmark{triangle}};
\addplot table [x=length, y={peak_mem}] {figures/dat/parity_check-transformer_encoder-speed-1024-full.dat} node[pos=1, opacity=1.0, scale=1.5, solid] {\pgfuseplotmark{square}};
\end{axis}
\end{tikzpicture}

%% file: figures/tikz/profiling_rtx4070.tex
\begin{tikzpicture}
\begin{axis}[
    axis x line=middle,
    axis y line=left,
    xlabel={Sequence Length $T$},
    ylabel={$\tau_p$ (ms $\downarrow$)},
    xmin=0, xmax=450,
    ymin=0, ymax=550,
    ylabel style={color=RoyalBlue},
    xtick align=inside,
    ytick={100, 200, 300, 400, 500},
    ytick style={color=RoyalBlue},
    yticklabel style={color=RoyalBlue},
    ylabel style={yshift=-5pt},
    cycle list name=primary,
    extra y ticks = {500},
    extra y tick style = {
        grid = major,
        major grid style = {dashed},
        tick style = {draw=none}
    },
    extra y tick label = \empty,
    width=4cm,
    height=3cm,
    scale only axis,
    font=\scriptsize
]
\addplot table [x=length, y=time] {figures/dat/parity_check-ntr-speed-1024.dat} node[pos=1, opacity=1.0, scale=1.5, solid] {\pgfuseplotmark{triangle}};
\addplot table [x=length, y=time] {figures/dat/parity_check-lstm-speed-1024.dat} node[pos=1, opacity=1.0, scale=1.5, solid] {\pgfuseplotmark{triangle}};
\addplot table [x=length, y=time] {figures/dat/parity_check-transformer_encoder-speed-1024.dat} node[pos=1, opacity=1.0, scale=1.5, solid] {\pgfuseplotmark{square}};
\end{axis}

\begin{axis}[
    at={(0,0)},
    anchor=origin,
    axis x line=none,
    axis y line=left,
    ylabel={Memory (GB $\downarrow$)},
    xmin=0, xmax=450,
    ymin=0, ymax=10.21,
    ylabel style={color=BrickRed},
    ytick style={color=BrickRed},
    yticklabel style={color=BrickRed},
    ylabel style={yshift=-5pt},
    cycle list name=primary,
    y dir=reverse,
    extra y ticks = {9.285},
    extra y tick style = {
        grid = major,
        tick style = {draw=none}
    },
    extra y tick label = \empty,
    width=4cm,
    height=3cm,
    scale only axis,
    font=\scriptsize
]
\addplot table [x=length, y={peak_mem}] {figures/dat/parity_check-ntr-speed-1024.dat} node[pos=1, opacity=1.0, scale=1.5, solid] {\pgfuseplotmark{triangle}};
\addplot table [x=length, y={peak_mem}] {figures/dat/parity_check-lstm-speed-1024.dat} node[pos=1, opacity=1.0, scale=1.5, solid] {\pgfuseplotmark{triangle}};
\addplot table [x=length, y={peak_mem}] {figures/dat/parity_check-transformer_encoder-speed-1024.dat} node[pos=1, opacity=1.0, scale=1.5, solid] {\pgfuseplotmark{square}};
\end{axis}
\end{tikzpicture}

%% file: figures/tabular/big_table_ablation.tex
\renewcommand{\arraystretch}{1.2}
\begin{tabular}{|cc|c|c>{\columncolor{gray!10}}cc|ccc|}
    \hline
    \multirow{3}{*}{Level} & \multirow{3}{*}{Task} & \multicolumn{7}{c|}{Score (\%)} \\
    \hhline{~~-------}
    & & \makecell{Parameter\\Matched} & \multicolumn{6}{c|}{Processing Depth}\\
    \hhline{~~-------}
    & & PR-LSTM & \multicolumn{3}{c|}{PR-LSTM} & \multicolumn{3}{c|}{PR-RNN} \\
    \noalign{\vskip -\arrayrulewidth}
    \hhline{~~-------}
    & & $d_h=137$ & $0R$ & $1R$ & $2R$ & $0R$ & $1R$ & $2R$ \\
    \hline
    \multirow{4}{*}{R} & %
    Even Pairs & \textbf{100} & \textbf{100} & \textbf{100} & \textbf{100} & \textbf{100} & \textbf{100} & \textbf{100}  \\
    & \makecell{Modular Arithmetic\\(Simple)} & \textbf{98.3} & 80.2 & \textbf{99.5} & \textbf{96.3} & \textbf{96.7} & 20.1 & 20.0 \\
    & Parity Check & \textbf{100} & \textbf{100} & \textbf{100} & \textbf{100} & \textbf{100} & \textbf{95.4} & 88.0 \\
    & Cycle Navigation & \textbf{97.4} & 48.4 & \textbf{94.4} & \textbf{100} & \textbf{100} & \textbf{99.9} & \textbf{100} \\
    \hline
    \multirow{4}{*}{DCF} & %
    Stack Manipulation & 55.5 & 54.4 & 55.6 & 55.8 & 51.9 & 52.1 & 50.2 \\
    & Reverse String & 54.7 & 55.9 & 55.5 & 54.8 & 54.9 & 50.1 & 50.0 \\
    & Modular Arithmetic & 35.8 & 33.0 & 35.6 & 36.7 & 22.0 & 25.1 & 25.3 \\
    & Solve Equation & 34.8 & 22.8 & 32.9 & 32.1 & 22.2 & 20.0 & 19.9 \\
    \hline
    \multirow{4}{*}{CS} & %
    Duplicate String & 52.9 & 53.4 & 53.5 & 54.0 & 50.9 & 51.1 & 50.1 \\
    & Missing Duplicate & \textbf{100} & 54.0 & \textbf{100} & \textbf{100} & 52.8 & \textbf{99.8} & \textbf{97.6} \\
    & Odds First & 53.4 & 53.5 & 54.3 & 54.2 & 51.1 & 50.7 & 50.0 \\
    & Binary Addition & 51.7 & 50.8 & 51.2 & 51.9 & 48.9 & 48.4 & 48.0 \\
    & Binary Multiplication & 52.6 & 51.8 & 52.2 & 52.2 & 49.9 & 46.7 & 49.7 \\
    & Compute Sqrt & 55.7 & 56.2 & 56.8 & 57.0 & 53.4 & 50.2 & 50.2 \\
    & Bucket Sort & \textbf{99.4} & \textbf{99.2} & \textbf{98.3} & \textbf{99.5} & 29.8 & 68.7 & 85.2 \\
    \hline
    \multicolumn{2}{|c|}{Tasks Solved} & 6 & 3 & 6 & 6 & 4 & 4 & 3 \\
    \hline
\end{tabular}

%% file: figures/tikz/time_efficiency_full.tex
    \begin{tikzpicture}
        \begin{groupplot}[
            group style={
                group size=3 by 5,
                horizontal sep=2cm,
                vertical sep=2cm,
            },
            title style={
                font=\small\bf,
            },
            enlarge x limits=0.05,
            enlarge y limits=0.05,
            cycle list name=primary,
            every axis plot/.style={opacity=0.75}
        ]
    
        \nextgroupplot[title={Even Pairs}];
        \addplot+[mark=*, mark indices={201}] table [x=time, y=loss] {figures/dat/even_pairs-lscm-training.dat};
        \addplot+[mark=*, mark indices={201}] table [x=time, y=loss] {figures/dat/even_pairs-lstm-training.dat};
        \addplot+[mark=*, mark indices={201}] table [x=time, y=loss] {figures/dat/even_pairs-transformer_encoder-training.dat};
        \nextgroupplot[title={Modular Arithmetic (Simple)}];
        \addplot+[mark=*, mark indices={201}] table [x=time, y=loss] {figures/dat/modular_arithmetic-lscm-training.dat};
        \addplot+[mark=*, mark indices={201}] table [x=time, y=loss] {figures/dat/modular_arithmetic-lstm-training.dat};
        \addplot+[mark=*, mark indices={201}] table [x=time, y=loss] {figures/dat/modular_arithmetic-transformer_encoder-training.dat};
        \nextgroupplot[title={Parity Check}]
        \addplot+[mark=*, mark indices={201}] table [x=time, y=loss] {figures/dat/parity_check-lscm-training.dat};
        \addplot+[mark=*, mark indices={201}] table [x=time, y=loss] {figures/dat/parity_check-lstm-training.dat};
        \addplot+[mark=*, mark indices={201}] table [x=time, y=loss] {figures/dat/parity_check-transformer_encoder-training.dat};
        \nextgroupplot[title={Cycle Navigation}]
        \addplot+[mark=*, mark indices={201}] table [x=time, y=loss] {figures/dat/cycle_navigation-lscm-training.dat};
        \addplot+[mark=*, mark indices={201}] table [x=time, y=loss] {figures/dat/cycle_navigation-lstm-training.dat};
        \addplot+[mark=*, mark indices={201}] table [x=time, y=loss] {figures/dat/cycle_navigation-transformer_encoder-training.dat};
        \nextgroupplot[title={Stack Manipulation}]
        \addplot+[mark=*, mark indices={201}] table [x=time, y=loss] {figures/dat/stack_manipulation-lscm-training.dat};
        \addplot+[mark=*, mark indices={201}] table [x=time, y=loss] {figures/dat/stack_manipulation-lstm-training.dat};
        \addplot+[mark=*, mark indices={201}] table [x=time, y=loss] {figures/dat/stack_manipulation-transformer_encoder-training.dat};
        \nextgroupplot[title={Reverse String}]
        \addplot+[mark=*, mark indices={201}] table [x=time, y=loss] {figures/dat/reverse_string-lscm-training.dat};
        \addplot+[mark=*, mark indices={201}] table [x=time, y=loss] {figures/dat/reverse_string-lstm-training.dat};
        \addplot+[mark=*, mark indices={201}] table [x=time, y=loss] {figures/dat/reverse_string-transformer_encoder-training.dat};
        \nextgroupplot[title={Modular Arithmetic}, ylabel={Training loss ($\downarrow$)}]
        \addplot+[mark=*, mark indices={201}] table [x=time, y=loss] {figures/dat/modular_arithmetic_brackets-lscm-training.dat};
        \addplot+[mark=*, mark indices={201}] table [x=time, y=loss] {figures/dat/modular_arithmetic_brackets-lstm-training.dat};
        \addplot+[mark=*, mark indices={201}] table [x=time, y=loss] {figures/dat/modular_arithmetic_brackets-transformer_encoder-training.dat};
        \nextgroupplot[title={Solve Equation}]
        \addplot+[mark=*, mark indices={201}] table [x=time, y=loss] {figures/dat/solve_equation-lscm-training.dat};
        \addplot+[mark=*, mark indices={201}] table [x=time, y=loss] {figures/dat/solve_equation-lstm-training.dat};
        \addplot+[mark=*, mark indices={201}] table [x=time, y=loss] {figures/dat/solve_equation-transformer_encoder-training.dat};
        \nextgroupplot[title={Duplicate String}]
        \addplot+[mark=*, mark indices={201}] table [x=time, y=loss] {figures/dat/duplicate_string-lscm-training.dat};
        \addplot+[mark=*, mark indices={201}] table [x=time, y=loss] {figures/dat/duplicate_string-lstm-training.dat};
        \addplot+[mark=*, mark indices={201}] table [x=time, y=loss] {figures/dat/duplicate_string-transformer_encoder-training.dat};
        \nextgroupplot[title={Missing Duplicate}]
        \addplot+[mark=*, mark indices={201}] table [x=time, y=loss] {figures/dat/missing_duplicate_string-lscm-training.dat};
        \addplot+[mark=*, mark indices={201}] table [x=time, y=loss] {figures/dat/missing_duplicate_string-lstm-training.dat};
        \addplot+[mark=*, mark indices={201}] table [x=time, y=loss] {figures/dat/missing_duplicate_string-transformer_encoder-training.dat};
        \nextgroupplot[title={Odds First}]
        \addplot+[mark=*, mark indices={201}] table [x=time, y=loss] {figures/dat/odds_first-lscm-training.dat};
        \addplot+[mark=*, mark indices={201}] table [x=time, y=loss] {figures/dat/odds_first-lstm-training.dat};
        \addplot+[mark=*, mark indices={201}] table [x=time, y=loss] {figures/dat/odds_first-transformer_encoder-training.dat};
        \nextgroupplot[title={Binary Addition}]
        \addplot+[mark=*, mark indices={201}] table [x=time, y=loss] {figures/dat/binary_addition-lscm-training.dat};
        \addplot+[mark=*, mark indices={201}] table [x=time, y=loss] {figures/dat/binary_addition-lstm-training.dat};
        \addplot+[mark=*, mark indices={201}] table [x=time, y=loss] {figures/dat/binary_addition-transformer_encoder-training.dat};
        \nextgroupplot[title={Binary Multiplication}]
        \addplot+[mark=*, mark indices={201}] table [x=time, y=loss] {figures/dat/binary_multiplication-lscm-training.dat};
        \addplot+[mark=*, mark indices={201}] table [x=time, y=loss] {figures/dat/binary_multiplication-lstm-training.dat};
        \addplot+[mark=*, mark indices={201}] table [x=time, y=loss] {figures/dat/binary_multiplication-transformer_encoder-training.dat};
        \nextgroupplot[title={Compute Sqrt}, xlabel={Wall-clock training time}]
        \addplot+[mark=*, mark indices={201}] table [x=time, y=loss] {figures/dat/compute_sqrt-lscm-training.dat};
        \addplot+[mark=*, mark indices={201}] table [x=time, y=loss] {figures/dat/compute_sqrt-lstm-training.dat};
        \addplot+[mark=*, mark indices={201}] table [x=time, y=loss] {figures/dat/compute_sqrt-transformer_encoder-training.dat};
        \nextgroupplot[title={Bucket Sort}]
        \addplot+[mark=*, mark indices={201}] table [x=time, y=loss] {figures/dat/bucket_sort-lscm-training.dat};
        \addplot+[mark=*, mark indices={201}] table [x=time, y=loss] {figures/dat/bucket_sort-lstm-training.dat};
        \addplot+[mark=*, mark indices={201}] table [x=time, y=loss] {figures/dat/bucket_sort-transformer_encoder-training.dat};
        \end{groupplot}
    \end{tikzpicture}